\documentclass[11pt,letterpaper]{article}

\usepackage{authblk}
\usepackage{conll2018}
\usepackage{pslatex}
\usepackage[english]{babel}
\usepackage[utf8]{inputenc}
\usepackage{amsmath}
\usepackage{bm}
\usepackage{graphicx}
\usepackage{tikz}
\usepackage{xcolor}
\usepackage{url}
\usepackage{rotating}
\usepackage{natbib}
\usepackage{amssymb}
\usepackage{linguex}
\usepackage{xspace}
\usepackage{subfig}

\newcommand{\key}[1]{\textbf{#1}}
\newcommand{\soft}[1]{}
\newcommand{\nopreview}[1]{}

\newcommand{\gulordavarnn}{GRNN\xspace}
\newcommand{\googlernn}{JRNN\xspace}
\newcommand{\jprnn}{JPRNN\xspace}

\aclfinalcopy

\title{RNNs as psycholinguistic subjects: Syntactic state and grammatical dependency}
\author[1]{\textbf{Richard Futrell}}
\author[2]{\textbf{Ethan Wilcox}}
\author[3,4]{\textbf{Takashi Morita}}
\author[5]{\textbf{Roger Levy}}

\affil[1]{Department of Language Science, UC Irvine, \tt{rfutrell@uci.edu}}

\affil[2]{Department of Linguistics, Harvard University, \tt{wilcoxeg@g.harvard.edu}}
\affil[3]{Primate Research Institute, Kyoto University, \tt{tmorita@alum.mit.edu}}
\affil[4]{Department of Linguistics and Philosophy, MIT}
\affil[5]{Department of Brain and Cognitive Sciences, MIT, \tt{rplevy@mit.edu}}

\date{}

\begin{document}

\maketitle


\begin{abstract} 
Recurrent neural networks (RNNs) are the state of the art in sequence modeling for natural language. However, it remains poorly understood what grammatical characteristics of natural language they implicitly learn and represent as a consequence of optimizing the language modeling objective.  Here we deploy the methods of controlled psycholinguistic experimentation to shed light on to what extent RNN behavior reflects incremental syntactic state and grammatical dependency representations known to characterize human linguistic behavior.  We broadly test two publicly available long short-term memory (LSTM) English sequence models, and learn and test a new Japanese LSTM. We demonstrate that these models represent and maintain incremental syntactic state, but that they do not always generalize in the same way as humans. Furthermore, none of our models learn the appropriate grammatical dependency configurations licensing reflexive pronouns or negative polarity items.  

\end{abstract}

\section{Introduction}
Progress in natural language processing has recently come from deriving sentence representations using recurrent neural networks (RNNs) \citep{elman1990finding,sutskever2014sequence,goldberg2017neural}. Yet while these networks have had great success, the nature of the representations they learn is unclear, which poses problems for interpretability, accountability, and controllability of NLP systems. More specifically, the success of RNNs has raised the question of whether and how these networks learn robust syntactic generalizations about natural language, which would enable robust performance even on data that differs from the peculiarities of the training set.

Here we build upon recent work studying RNN language models with the same techniques used to study language processing in the human mind: by examining their behavior on targeted sentences chosen to probe particular aspects of the learned representations. \citet{linzen2016assessing}, followed more recently by others \citep{bernardy2017using,enguehard2017exploring,gulordava2018colorless}, use an \key{agreement prediction task} \citep{bock1991broken} to study whether RNNs learn a hierarchical morphosyntactic dependency: for example, that \emph{The key to the cabinets\dots} can grammatically continue with \emph{was} but not with \emph{were}.  This dependency turns out to be learnable at human performance from a language modeling objective alone \citep{gulordava2018colorless}.  In the present work we extend this general approach to a wider range of grammatical phenomena.

We draw on the rich literature in human sentence processing to subject RNNs to much the same scrutiny as a human experimental subject in a psycholinguistic study might undergo: what linguistic knowledge does the subject's incremental processing behavior reflect?  We focus on two central types of knowledge that may be evident in processing: \textbf{syntactic state}, a representation of syntactic events that may have occurred, that are currently unfolding, and that are yet to come; and \textbf{grammatical dependency}, the set of conditions characterizing syntactically mediated relations among elements in a sentence.  We investigate syntactic state by studying RNNs' behavior under garden-path disambiguation, multiple center-embedding, and the maintenance of obligatory upcoming syntactic events.  For grammatical dependency, we extend the above recent studies of verb agreement by investigating the binding of reflexive pronouns and the licensing of negative polarity items.  Beyond helping characterize the representations of contemporary RNNs, this work may bear on classic learnability questions in acquisition: what grammatical knowledge can be learned from a childhood's or a lifetime's worth of string input by a flexible sequence model without a strong domain-specific inductive bias?


\section{General methods}

We investigate RNN behavior primarily by studying the \key{surprisal}, or log inverse probability, that an RNN assigns to each word in a sentence:
\begin{equation*}
S(x_i) = -\log_2 p(x_i|h_{i-1}),
\end{equation*}
where $x_i$ is the current word or character, $h_{i-1}$ is the RNN's hidden state before consuming $x_i$, the probability is calculated from the RNN's softmax activation, and the logarithm is taken in base 2, so that surprisal is measured in bits. 

A common practice in psycholinguistics is to study a measure of reaction time per word (for example reading time as measured by an eyetracker), as a measure of the word-by-word difficulty of online language processing. These reading times are often taken to reflect the extent to which humans expect certain words in context, and may be generally proportional to surprisal given the comprehender's probabilistic language model \citep{hale2001probabilistic,levy2008expectation,smith2013effect}. In this study, we take RNN surprisal as the analogue of human reading time, using it to probe the RNNs' expectations about what words will follow in certain contexts. While we are not interested in directly modeling human processing difficulty here, we note that there is a long tradition linking RNN performance to human language processing \citep{elman1990finding,christiansen1999toward,macdonald-christiansen:2002,frank2011insensitivity}.  

\subsection{Experimental methodology}

In each experiment presented below, we design a set of sentences such that the word-by-word surprisal values will show evidence for syntactic representations.\footnote{Our experiments and analyses were preregistered on \url{aspredicted.org}: blind preregistration codes \texttt{5vr6ze}, \texttt{f8yd86}, \texttt{vh82i7}, \texttt{pt3x3i}, \texttt{yt6pi4}.} We analyze by-word surprisal profiles for these sentences using regression analysis. 

Except where otherwise noted, all statistics are derived from linear mixed-effects models \citep{baayen2008mixed} with sum-coded fixed-effects predictors and by-item random intercepts, where the dependent variable is the summed surprisal across words within the region in question.  Random slopes and interactions are not necessary in these models to avoid anti-conservativity \citep{barr2013random} because we do not have repeated observations within any item/condition combination. This method allows us to factor out by-item variation in surprisal and focus on the contrasts between conditions.

\subsection{LSTMs tested}

We study the behavior of two LSTMs for English. First, the model presented in \citet{jozefowicz2016exploring} as ``BIG LSTM+CNN Inputs'', which we call ``\googlernn'', which was trained on the One Billion Word Benchmark \citep{chelba2013one} with two hidden layers of 8196 units each and CNN character embeddings as input. Second, we use the model described in the supplementary materials of \citet{gulordava2018colorless}, which we call ``\gulordavarnn'', trained on 90 million tokens of English Wikipedia with two hidden layers of 650 hidden units each. In Section~\ref{sec:japanese-npi}, we also study the behavior of an LSTM for Japanese (\jprnn), which is a character-based model \citep[cf.][]{Kim_et_al_2016_character-based_NN}%
\footnote{Our model does not have a convolutional layer, but rather only an embedding layer, due to the considerable size of the vocabulary.} %
with 650 hidden units, trained on 900,000 paragraphs (800,000 for training and 100,000 for validation) of Japanese Wikipedia. After 100 epochs, we obtained the best validation perplexity, at 12.67.  All LSTMs are trained on a pure language modeling objective. 

Our goal in examining these models is not to draw contrasts between them, since they are very similar in their architecture and performance (in terms of perplexity); rather our goal is to provide results from samples of state-of-the-art models. Future work may examine how our findings differ across model architectures and draw causal connections between model architectures and the ability to represent complex syntax.

\section{Garden path effects}
\label{sec:garden-path}

One of the major questions in human sentence processing has been how people represent the incremental parse of a sentence during online language comprehension. The major phenomenon that has been used to probe these representations is \key{garden path effects}. Garden path effects arise from local ambiguities, where a context leads a person to believe one parse is likely, but then a disambiguating word forces her to drastically revise her beliefs. In effect, the comprehender is ``led down the garden path'' by a locally likely but ultimately incorrect parse \citep{bever1970cognitive}. 

In psycholinguistics, garden path effects have been studied in order to answer questions like: do humans represent multiple possible parses in parallel, ranked by probability, or do they only represent the single most likely parse? What information affects the parse tree a person will consider most likely given a locally ambiguous context \citep{macdonald1994lexical}? For RNNs, these methods can be used to answer the questions above, and also: what properties of the input lead to syntactic representations that are more or less accurate?

Garden-pathing in RNNs has very recently been demonstrated by \citet{vanschijndel2018modeling}, albeit over only a short (two-word) stretch of ambiguity-maintaining material.  Here we investigate a garden path previously unstudied in RNNs, induced by the classic Main Verb/Reduced Relative (MV/RR) ambiguity, in which a word is locally ambiguous between being the main verb of a sentence or introducing a reduced relative clause, and that ambiguity is maintained over a longer stretch of material:

{
\setlength{\Exlabelwidth}{0.5em}
\setlength{\Exlabelsep}{0.5em}
\setlength{\SubExleftmargin}{0.7em}
\small
\ex. \label{ex:mv-rr}
\a. The woman brought the sandwich from the kitchen tripped on the carpet. [\textsc{reduced}, \textsc{ambig}uous]\label{ex:mv-rr-reduced-ambig}
\b. The woman given the sandwich from the kitchen tripped on the carpet. [\textsc{reduced}, \textsc{unambig}uous]\label{ex:mv-rr-reduced-unambig}
\c. The woman who was brought the sandwich from the kitchen tripped on the carpet. [\textsc{unreduced}, \textsc{ambig}]
\d. The woman who was given the sandwich from the kitchen tripped on the carpet. [\textsc{unreduced}, \textsc{unambig}] \label{ex:mv-rr-unreduced-unambig} 

}

In Example~\ref{ex:mv-rr-reduced-ambig}, the phrase ``brought the sandwich from the kitchen'' is initially analyzed as a main verb phrase, but upon reaching the verb ``tripped''---the \key{disambiguator}---the reader must re-analyze it as a relative clause. In these examples, there are two possible cues that the first verb is introducing a reduced relative clause (reduced RC): either (i) it can be preceded with ``who was'' (the RC is \key{unreduced}) or (ii) it is ambiguously a past participle or a past tense verb, such as ``brought'' rather than ``given''---that is, it is \key{ambiguous}. The garden path effect should only arise in the critical condition where the relative clause is reduced and the verb is ambiguous. 

For our purposes, the key dependent variable for MV/RR sentences is the surprisal at the disambiguator. If the surprisal at the disambiguator is higher in the critical condition than in the other conditions, this indicates that the network had a preferred syntactic analysis for the previous material which did not lead it to expect the disambiguator. 

The surprisal in the region \emph{following} the disambiguator is also interesting. If it is the same across conditions, that indicates that the network successfully revised its syntactic analysis. If it is high in the critical condition, then the network did not recover from the garden path event: that is, either the disambiguator did not provide sufficient information to cause the network to revise its syntactic analysis, or it caused the network to enter a confused state from which only poor predictions can be made. 

We present results from two manipulations of the MV/RR ambiguity. First, we present an experiment manipulating verb ambiguity, as presented in \ref{ex:mv-rr-reduced-ambig}--\ref{ex:mv-rr-unreduced-unambig}. Second, we present an experiment where the subject NP provides cues about whether a following ambiguous verb should be interpreted as a main verb or a reduced RC.

\subsection{Verbform ambiguity and RC reduction}
\label{sec:mv-rr-verb-ambiguity}

\begin{figure*}
\centering
\includegraphics[width=0.9\textwidth]{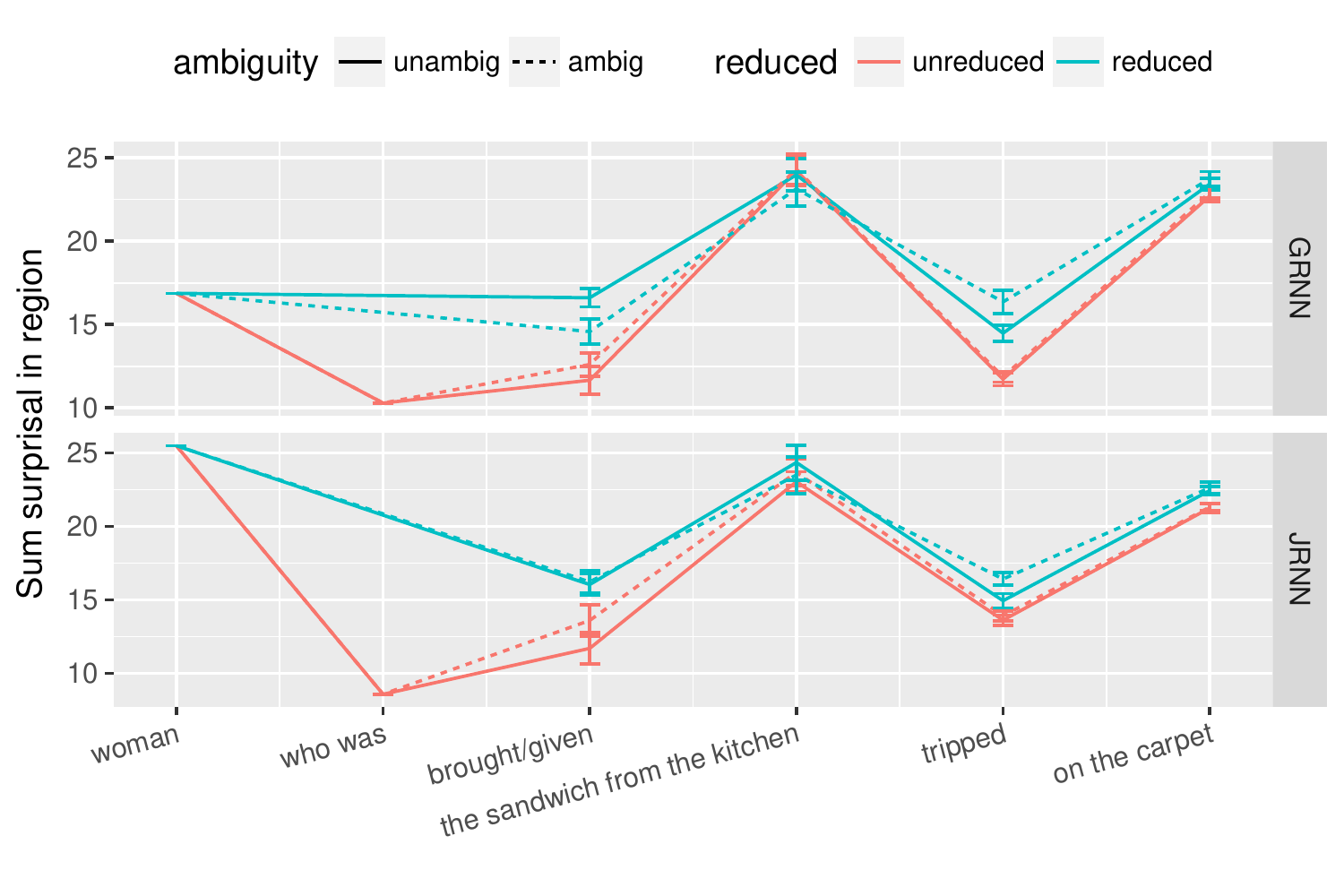} 
\vspace{-0.5cm}
\caption{Surprisal by sentence region in garden-pathing study on verbform ambiguity and RC reduction.  In this and all other figures, unless otherwise noted, \key{error bars} represent 95\% confidence intervals of the \emph{contrasts between conditions}, computed from the standard error of the by-item and by-condition mean surprisals after subtracting out the by-item means \citep{masson2003using}. In cases where there is no contrast in surprisal across conditions, these intervals should be zero.} 
\label{fig:mv-rr-surprisal}
\end{figure*}

Figure~\ref{fig:mv-rr-surprisal} shows surprisals for each RNN in each sentence region for 29 items we constructed following the template of \ref{ex:mv-rr}.  Surprisals are summed over words in the region, giving the total unexpectedness of the phrase, and then averaged across items. 
At the critical main-clause verb, surprisal is superadditively highest in the reduced ambiguous condition (the dotted blue line; a positive interaction between the reduced and ambiguous conditions is significant in both models at $p<0.001$), the key predicted human-like garden-path disambiguation effect. Within the reduced conditions (represented by blue lines), surprisal is lower when the participial verbform was unambiguous than when it was ambiguous ( $p<0.001$ in \googlernn and $p<0.01$ in \gulordavarnn), demonstrating that the models have learned the distinctive syntactic behavior of participial verbs.  But strikingly, even when the participial verbform is unambiguous, surprisal is still higher when the RC is reduced than when it is unreduced (compare the red and blue solid lines; $p<0.001$ in both models), suggesting that the models have not fully representationally separated participial verbs from finite verbs. Apparently, the network treats an unambiguous participial verb as only a noisy cue to the presence of an RC. 

Post-disambiguation, surprisals are higher in the unreduced conditions than in the reduced conditions (both models $p<0.001$), suggesting that the models may not fully recover to a clean syntactic state following garden-path disambiguation.  

\subsection{Subject animacy}
\label{sec:mv-rr-animacy}

Syntactic garden-pathing in humans has been demonstrated to be sensitive to fine-grained lexical and semantic cues, such as the animacy of the NP subject in the case of MV/RR garden-pathing (\citealp{trueswell1994semantic}, though see \citealp{ferreira1986independence,clifton-etal:2003jml} for controversy regarding time-course).  Are RNNs similarly sensitive? We examined this question with 30 items on the \citeauthor{trueswell1994semantic} template of~\ref{ex:mv-rr-animacy} below:

{\small
\ex. \label{ex:mv-rr-animacy}
\a. The witness examined by the lawyer turned out to be unreliable. [\textsc{reduced}, \textsc{animate}]\label{ex:mv-rr-animacy-animate-reduced}
\b. The evidence examined by the lawyer turned out to be unreliable. [\textsc{reduced}, \textsc{inanimate}]\label{ex:mv-rr-animacy-inanimate-reduced}
\c. The witness that was examined by the lawyer turned out to be unreliable. [\textsc{unreduced}, \textsc{animate}]
\d. The evidence that was examined by the lawyer turned out to be unreliable. [\textsc{unreduced}, \textsc{inanimate}]
\z.

}

\noindent
If RNNs have human-like sensitivity to the fine-grained covariance of syntactic structure and lexico-semantic information, then surprisal should be superadditively higher in the reduced/animate condition~\ref{ex:mv-rr-animacy-animate-reduced}.  The strongest evidence for such a contingency would be if this effect shows up at the RC-internal \emph{by}-phrase, which has poor compatibility with an active-voice analysis of the preceding verbform.  

\begin{figure*}
\centering
\includegraphics[width=0.90\textwidth]{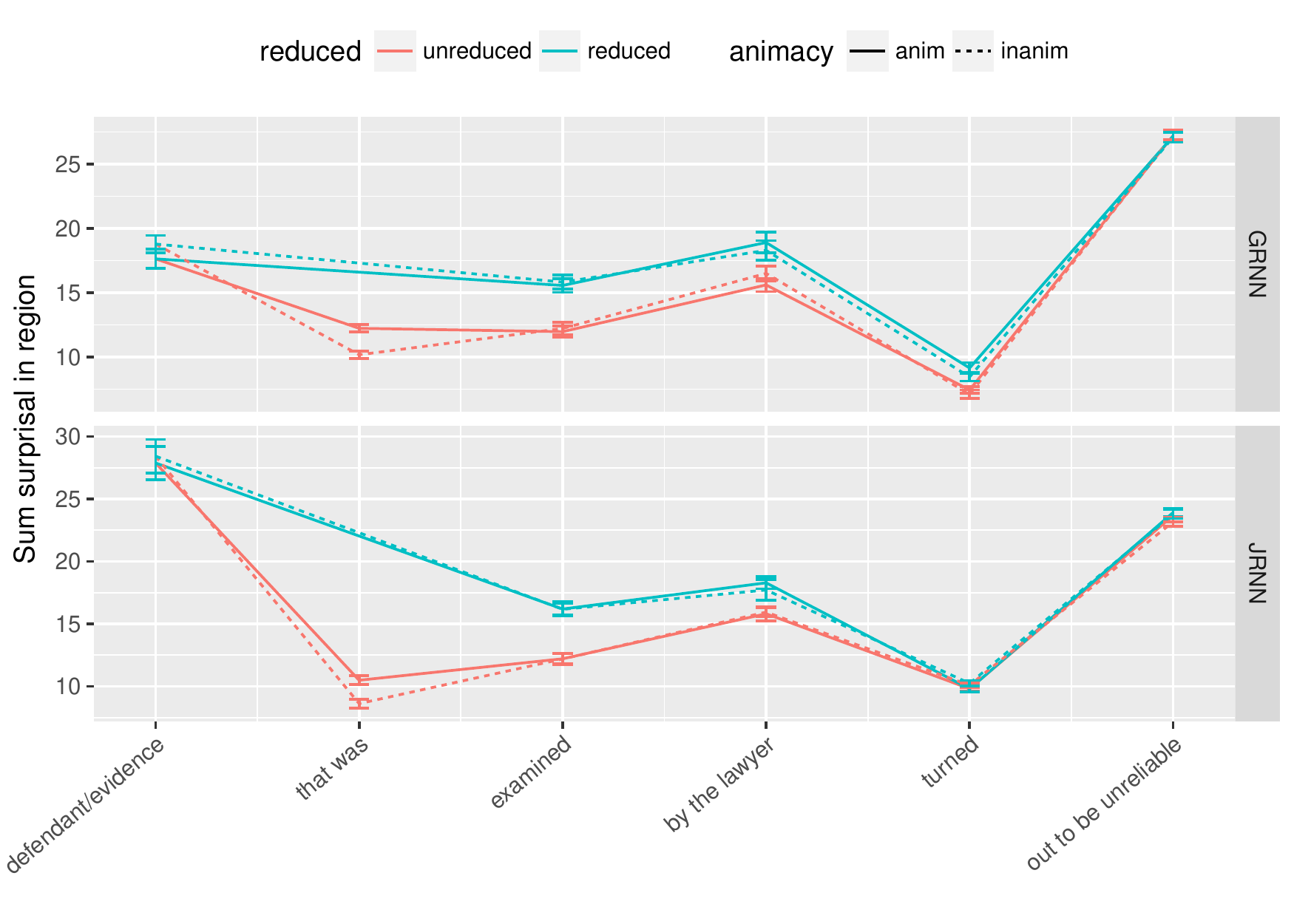}
\caption{Garden-pathing and subject animacy. The critical disambiguating region is ``by the lawyer''.}
\label{fig:animacy-garden-path}
\end{figure*}

Figure~\ref{fig:animacy-garden-path} shows surprisals by region, condition, and RNN, averaged across items.  At the \emph{by}-phrase there is a large main effect of RC reduction (compare the blue vs. red lines; both models $p<0.001$), and a small interaction between reduction and animacy in the predicted direction: with animate subject nouns, surprisal in this region is higher when the RC is reduced, but not when the RC is unreduced.  This interaction is significant in \gulordavarnn ($p=0.005$) but not in \googlernn ($p=0.1$). At the main verb, \gulordavarnn shows once again a large main effect of RC reduction ($p<0.001$), indicating that a main verb is still somewhat surprising following a reduced relative; \googlernn shows no such effect. The final region has high total surprisal because it is several words long. It shows a small but significant effect of RC reduction in \googlernn ($p=0.04$), and no other significant contrasts between conditions. 

Our result shows that \gulordavarnn can exploit fine-grained information about the covariance of lexical forms and syntactic structure in order to infer syntactic state. However, the presence of residual ``spillover'' effects at the main verb suggest that the network has not encountered sufficient evidence at the main verb to close the relative clause. 

\section{Obligatory upcoming syntactic events}
\label{sec:obligatory-upcoming}

Garden-path disambiguation effects are diagnostic of cases where a syntactic state is weighted strongly \emph{against} a syntactic event at a particular moment in incremental processing.  Other grammatical contexts create an obligation \emph{for} a certain type of syntactic event in the future.  In an incremental processing system such as a human or an RNN, this obligation must be maintained for an indefinitely long time.  It is trivial to do so in rule-based processing architectures with a stack, and sentence processing research clearly demonstrates that humans maintain such expectations in syntactic processing \citep{staub-clifton:2006,lau-etal:2006,levy-fedorenko-breen-gibson:2012cognition}, but it remains unclear whether RNNs learn to use their memory this way in natural language sequence prediction. 

We tested two simple grammatical configurations inducing an obligatory upcoming syntactic event: object-extracted relative clauses and subordination.

\subsection{Relative clause completions}

A prefix such as the one in \ref{ex:orc-1} signals the onset of an object-extracted relative clause (ORC). A grammatical continuation of the prefix must include two verb phrases: one to finish the relative clause, and one to finish the main clause. Similarly, Example~\ref{ex:orc-2} signals the onset of two nested ORCs: a grammatical continuation must contain three verb phrases. Humans can reliably generate grammatical continuations for prefixes such as \ref{ex:orc-1}, but struggle with prefixes such as \ref{ex:orc-2} (\citealp{yngve1960model,miller1963finitary,gibson1999memory,vasishth2010shortterm,frank2016crosslinguistic,futrell2017noisycontext}; sample grammatical completions in italics):

{\small
\ex. \label{ex:orc}
\a. The author who the editor\dots \textit{disliked sent in the manuscript}. \label{ex:orc-1}
\b. The manuscript that the author who the editor\dots \textit{disliked sent in was of low quality}. \label{ex:orc-2}
\z.

}

We tested the LSTMs' ability to represent the requirement for two verb phrases in the first case and three verb phrases in the second case, using 20 prefixes on the template of \ref{ex:orc} based on materials from \citet{gibson1999memory}. We sampled 9 completions per prefix per condition per LSTM by recurrently sampling from the softmax distribution of following words up until the generation of the end-of-sequence symbol. We (the authors) then judged grammaticality of completions by hand. We judged grammaticality solely based on whether the network generated the right number of syntactically appropriate verb phrases, where a verb phrase was counted as appropriate if it matched its subject in number. Grammatical errors in irrelevant parts of the continuations were ignored. We ignored continuations where a judgment was impossible due to generation of an \texttt{<UNK>} token.\footnote{We carried out these judgments ourselves because they require some linguistic sophistication to identify the relevant verb phrases.}  Statistical analysis for this study was performed using a mixed logit model with fixed and by-item random effects of embedding depth, LSTM, and their interaction, to account for repeated measures.

\begin{figure}
  \centering
\includegraphics[width=0.37\textwidth]{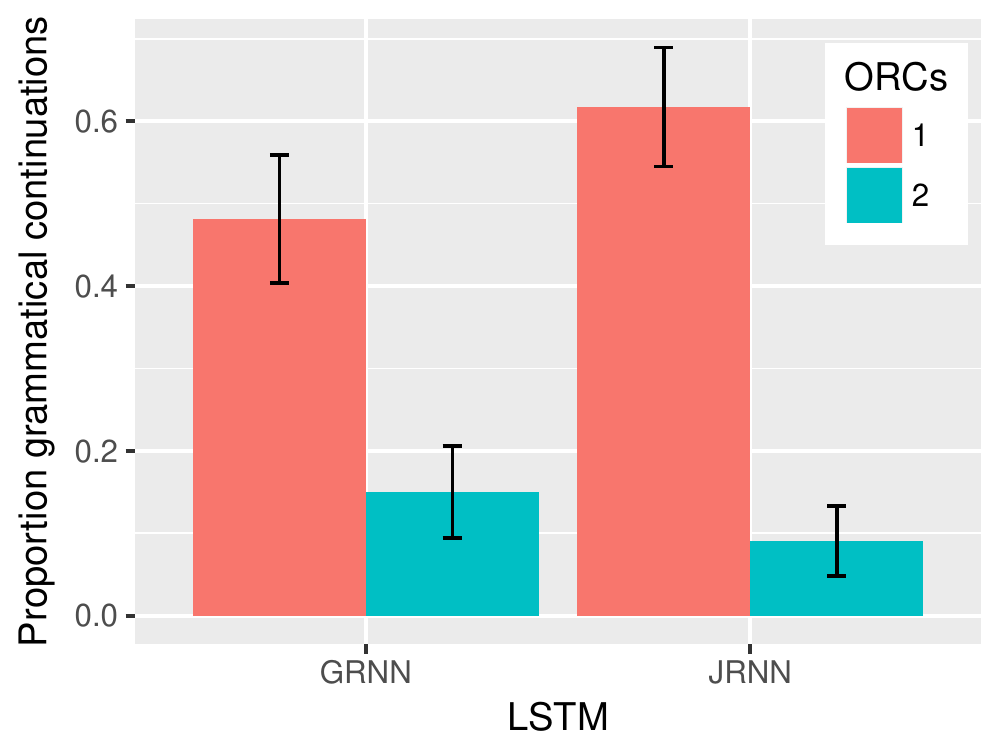}    
\caption{Proportions of grammatical prefix continuations, by LSTM and number of ORCs in prefix. Error bars are 95\% confidence intervals of by-item means.}
\label{fig:completions}
\end{figure}

Figure~\ref{fig:completions} shows the proportions of completions judged grammatical. Both \googlernn and \gulordavarnn generate a high proportion of grammatical continuations for one ORC, with \googlernn outperforming GRNN ($p<0.001$).  Neither network can reliably generate the required three verb phrases for a prefix with two nested ORCs, but GRNN suffers less than JRNN from the additional embedding, a significant interaction ($p<0.001$).

Relative clause completions seem to be a case where limitations in RNN performance mirror limitations in human performance. However, the networks have lower accuracy than human subjects across the board. Mechanical Turk subjects can complete single-ORC prefixes of this form grammatically with near 100\% accuracy and double-ORC prefixes with around 40--60\% accuracy (unpublished data: Gibson, p.c.). 

\subsection{Subordination}

If an English sentence begins with a subordinate clause, the expectation for the onset of the matrix clause must be maintained for however long the subordinate clause lasts \ref{ex:subordination-sub-matrix}.  Ending the sentence without a subordinate clause \ref{ex:subordination-sub-nomatrix} is surprising to humans and should be surprising to a human-like language model.  The strength of this syntactic obligation in a language model can be quantified by the size of the interaction effect between subordinator presence/absence and matrix-clause presence/absence on the joint surprisal of all post-subordinate clause material.\footnote{It is necessary to look at this $2 \times 2$ interaction rather than simply comparing \ref{ex:subordination-sub-matrix} and \ref{ex:subordination-sub-nomatrix} because we need to control for the surprisal of the following matrix clause. The logic of this interaction is similar to the ``1icensing interaction'' used to study filler--gap dependencies in \citet{wilcox2018what}.}

\begin{figure*}
\vspace{0pt}
\begin{minipage}{0.38\textwidth}
\includegraphics[width=\textwidth]{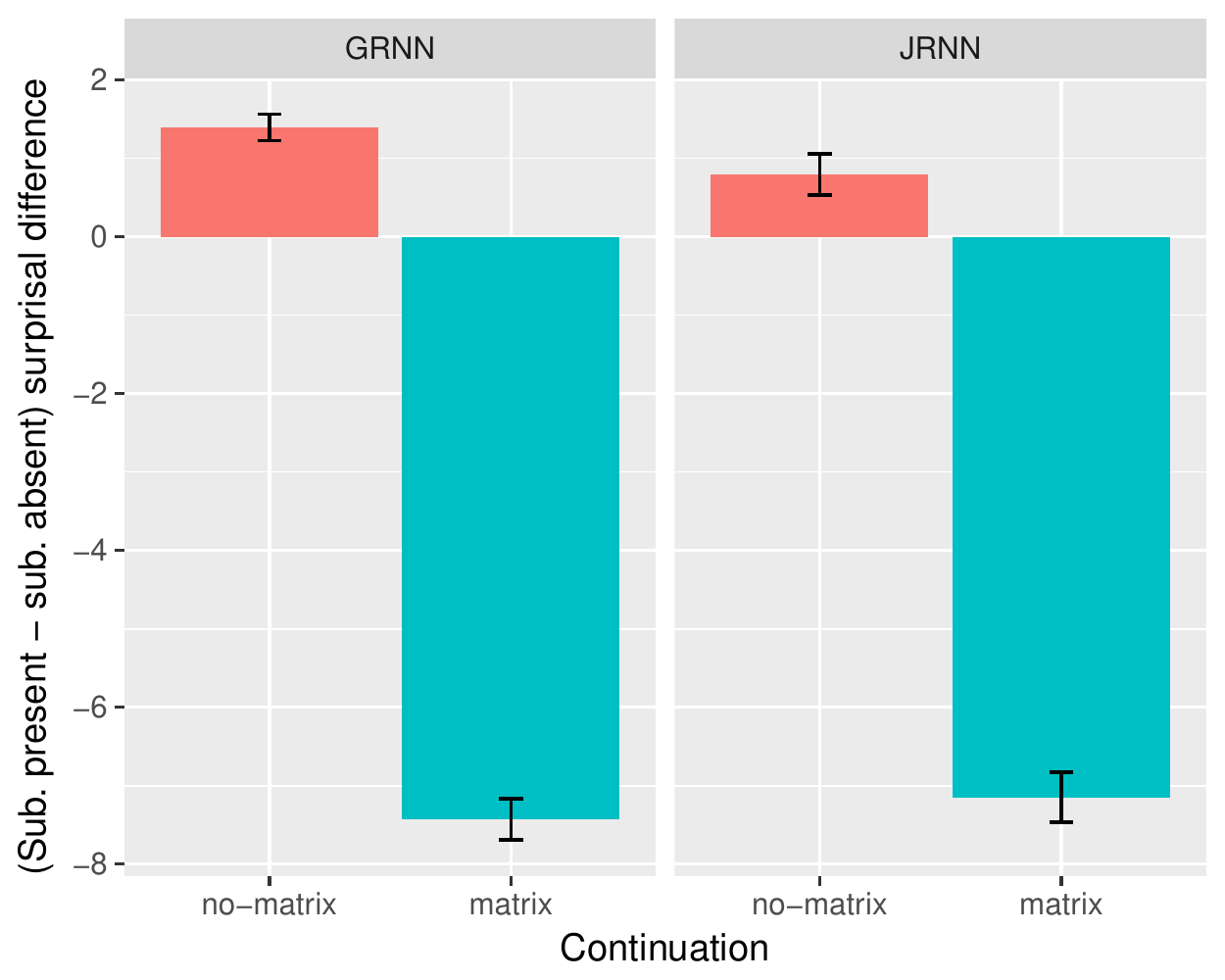}
\end{minipage}
\begin{minipage}{0.30\textwidth}
\includegraphics[width=\textwidth]{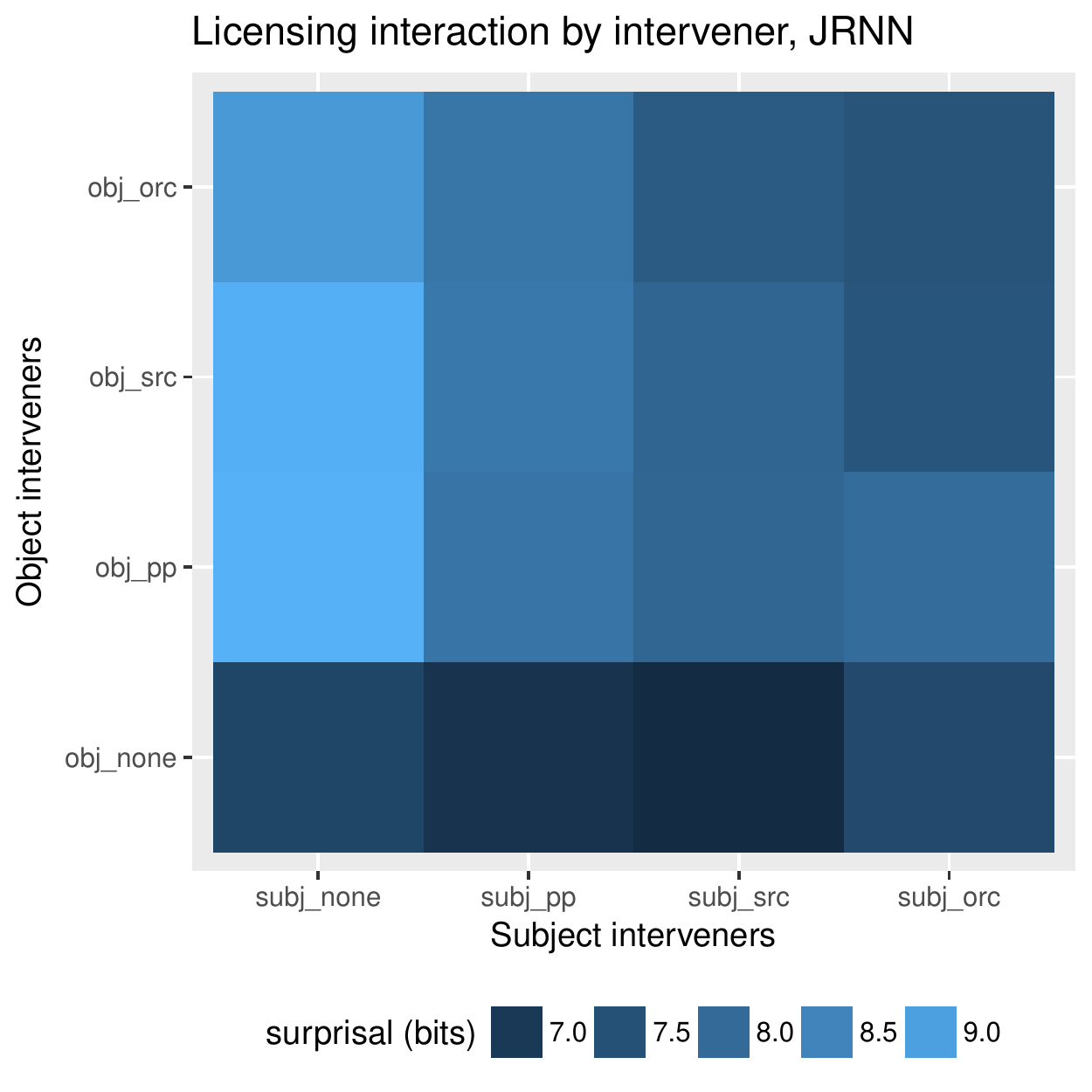}
\end{minipage}
\begin{minipage}{0.30\textwidth}
\includegraphics[width=\textwidth]{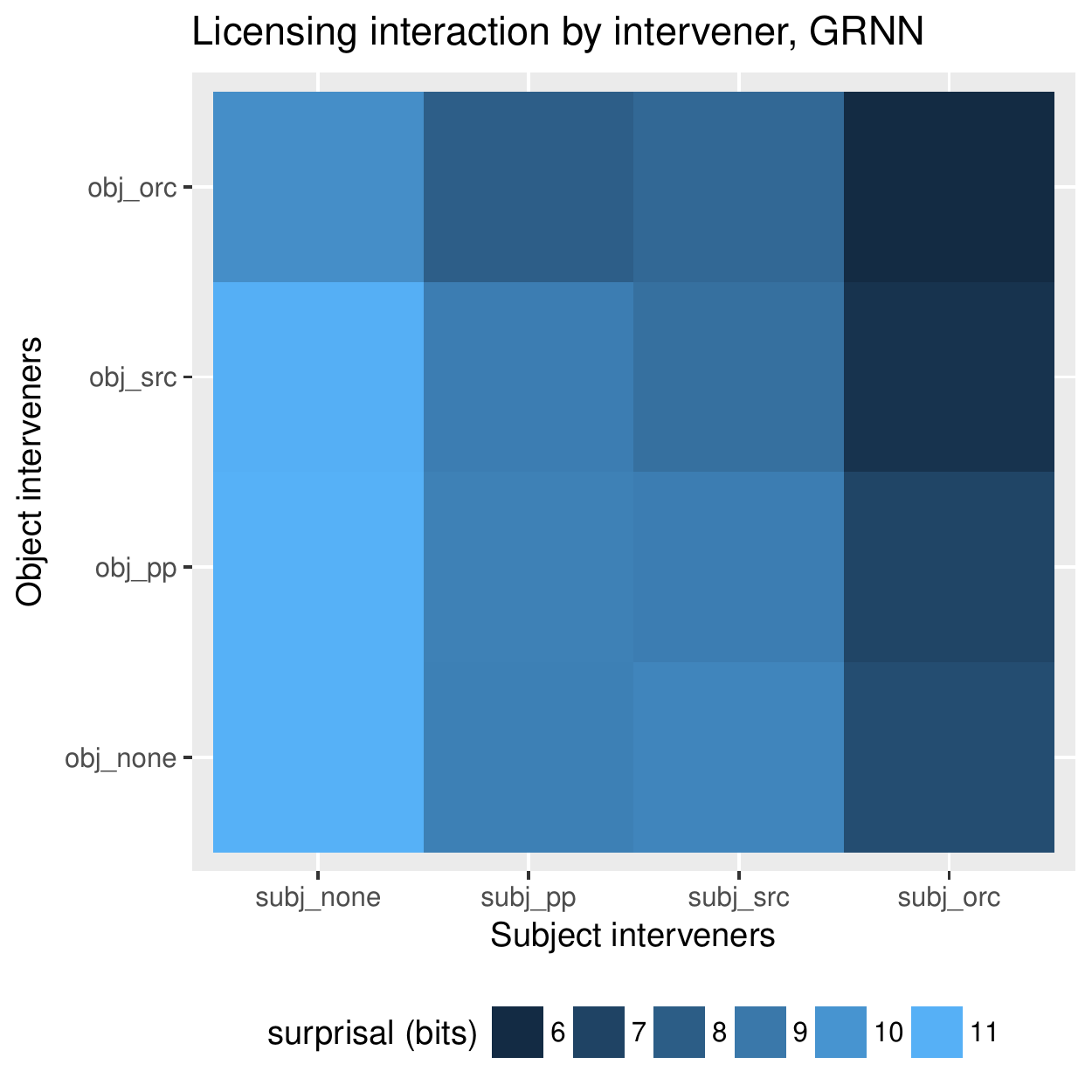}
\end{minipage}
\vspace{-0.15cm}
    \caption{Left panel: Effect of subordinator absence/presence on surprisal of continuations. Right two panels: Size of interaction effect between subordinator presence/absence and matrix clause presence/absence given various intervening elements in the subordinate clause.}
\label{fig:subordination-results}
\end{figure*}

{\small
\ex. \label{ex:subordination}
\a. As the doctor studied the textbook, the nurse walked into the office. \label{ex:subordination-sub-matrix}
\b. *As the doctor studied the textbook. \label{ex:subordination-sub-nomatrix}
\c. ?The doctor studied the textbook, the nurse walked into the office. \label{ex:subordination-nosub-matrix}
\d. The doctor studied the textbook. \label{ex:subordination-nosub-nomatrix}
\z.

}
We designed 23 items on the pattern of \ref{ex:subordination}; Figure~\ref{fig:subordination-results} (left and center panels) shows results from both LSTMs in terms of the difference in surprisal for matrix-clause and non-matrix-clause continuations depending on whether a subordinator is present or not. A positive effect in this figure indicates that a subordinator makes the continuation more surprising; a negative effect means the subordinator makes the continuation less surprising. We see a strong facilitative effect of the subordinator on matrix-clause continuations, and a somewhat weaker penalty on non-matrix-clause continuations in both models. As predicted, a negative interaction of matrix clause presence and subordinator presence is significant in both models ($p\text{s}<0.001$), and it is numerically much larger for GRNN.

We included a further manipulation in this study, optionally modifying each NP in the subordinate clause with a prepositional phrase or subject- or object-extracted relative clause, on the hypothesis that lengthening and increasing syntactic complexity of the subordinate clause might weaken the expectation for an upcoming matrix clause. 
Results can be seen in Figure~\ref{fig:subordination-results}, which shows the size of the interaction between presence of a subordinator and presence of a matrix clause (that is, the difference between the two bars for each model in the left panel of Figure~\ref{fig:subordination-results}). A positive interaction corresponds to a licensing relationship where the subordinator makes the matrix clause more likely and a premature ending less likely. \gulordavarnn exhibits a strong interaction when the intervening material is short and syntactically simple (Figure~\ref{fig:subordination-results}, left bottom), and the interaction gets progressively weaker as the intervening material becomes longer and more complex ($p<0.001$ for subject postmodifiers but not significant for object postmodifiers).
\googlernn has more complex behavior: object interveners actually make the matrix clause \emph{more} likely. Overall, in a linear regression, subject interveners decrease the size of the licensing interaction ($p=0.02$) and object interveners increase it ($p=0.005$). 

Taken together, the results of this study indicate that both LSTMs derive and maintain in memory an expectation for an upcoming matrix clause from a sentence-initial subordinator; that this expectation decays in the presence of complex intervening material; and that this expectation is stronger and exhibits more clearly understandable behavior in \gulordavarnn, even though that model is smaller and trained on less data.

\section{Reflexive pronoun binding and c-command}
\label{sec:refl-pron-bind}

Having investigated the representation and maintenance of syntactic state in both RNNs, we move on to examining their representation of grammatical dependencies that are defined with respect to syntactic state in human grammatical competence.  \citet{linzen2016assessing} and \citet{gulordava2018colorless} provide one such case, that of subject--verb agreement.  Here we extend the scope of these studies to \key{binding}, which characterizes the syntactic restrictions on pronouns and their antecedents \citep{chomsky1981lectures}. In English, a reflexive pronoun is subject to two constraints: (1) it must agree in gender and number with its antecedent, and (2) approximately, its antecedent must be the syntactically most local NP that \key{c-commands} it (\citealp{reinhart1981definite}; see \citealp{pollard1994headdriven} for a closely related characterization in a lexicalist syntactic framework).  Provided that an RNN learns NP stereotypical gender---a reasonable prospect given the results of \citet{caliskan-etal:2017semantics} and \citet{rudinger2018gender}---we can use reflexive pronoun surprisal to assess whether the model also learns the structure of the grammatical dependency that must characterize the relationship between a reflexive and its antecedent.

We chose 30 nouns referring to professions likely to have strong stereotypic gender, based on government statistical data \citep{uslaborforce}.  To assess whether each model learned this stereotypic gender, we constructed an item for each noun involving a simple transitive clause with a reflexive pronoun object of each gender:

{\small
\ex. \label{ex:stereotypical-gender}
\a. The hairdresser washed herself. [\textsc{match}]\label{ex:fem-fem}
\b. The hairdresser washed himself. [\textsc{mismatch}]\label{ex:fem-masc}
\c. The lumberjack cut himself. [\textsc{match}]\label{ex:masc-masc}
\d. The lumberjack cut herself. [\textsc{mismatch}]\label{ex:masc-fem}
\z.

}

\begin{figure}
\begin{minipage}{0.235\textwidth}
\includegraphics[width=\textwidth]{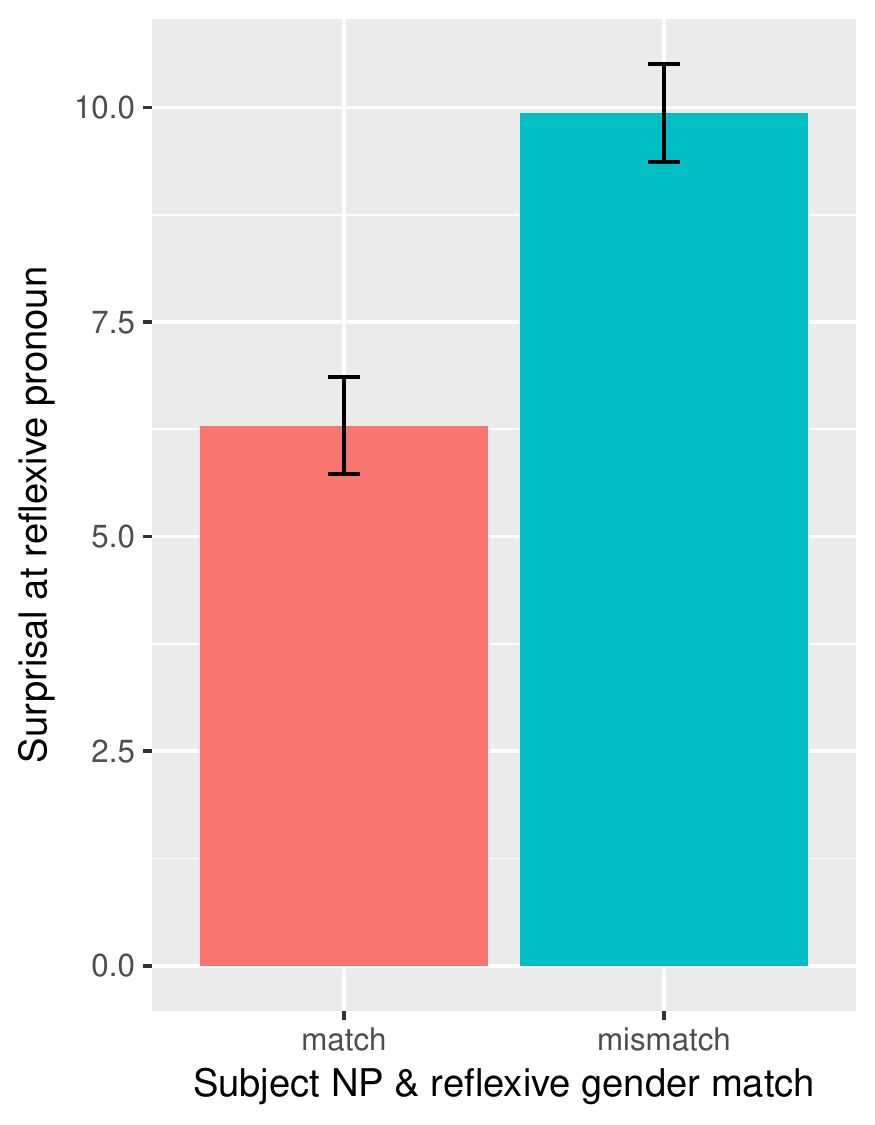}
\end{minipage}
\begin{minipage}{0.235\textwidth}
\includegraphics[width=\textwidth]{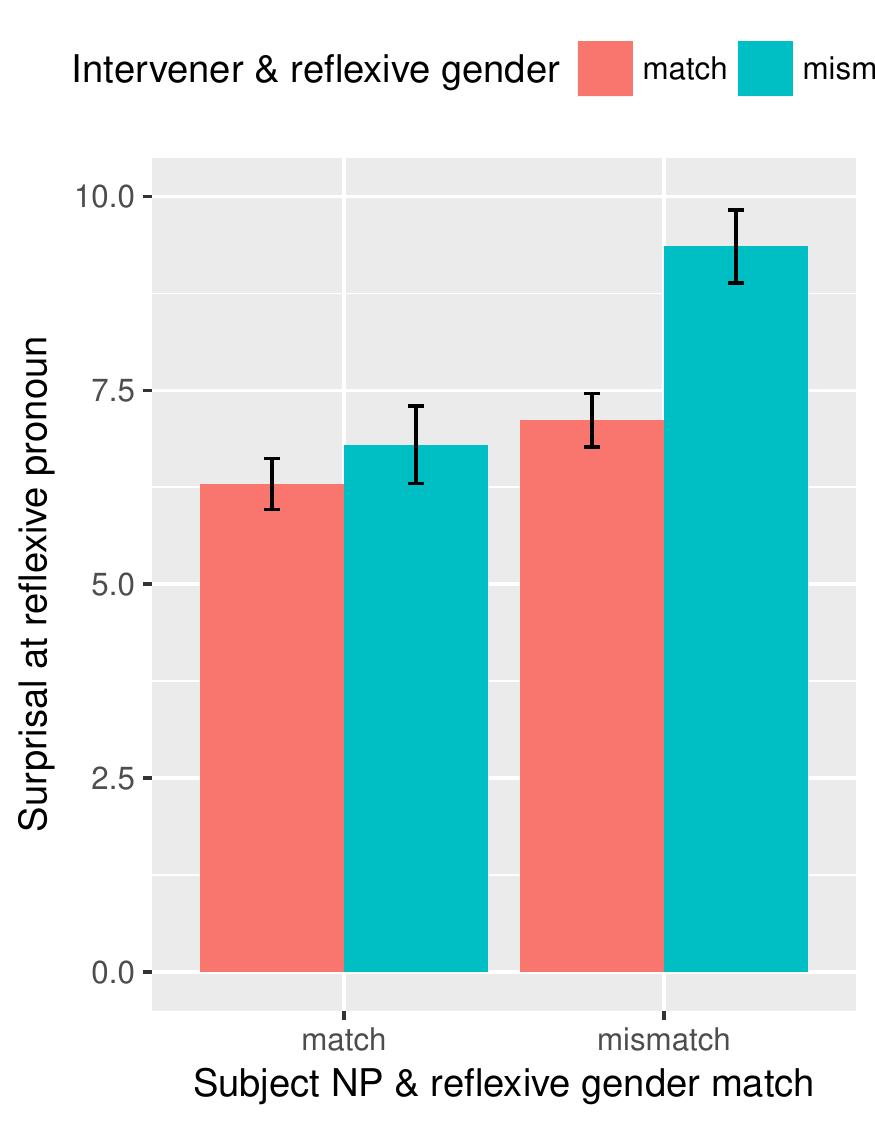}
\end{minipage}
\caption{\googlernn surprisal at reflexive pronouns. Left: no interveners as in example \ref{ex:stereotypical-gender}; right: with interveners; the bars correspond to examples \ref{ex:masc-masc-masc}, \ref{ex:masc-fem-masc}, \ref{ex:masc-masc-fem}, \ref{ex:masc-fem-fem} from left to right.}
\vspace{-0.3cm}
\label{fig:gender-surprisal}
\end{figure}

\noindent
For \googlernn, we find higher surprisal at reflexive pronouns mismatching the antecedent's stereotypical gender than at pronouns matching the antecedent's stereotypical gender (Figure~\ref{fig:gender-surprisal}, left panel, $p<0.001$).  We did not find a reliable effect of stereotypical gender for \gulordavarnn (not depicted), so we do not examine it further in this section.

Next, we tested whether \googlernn's probabilistic dependency between preceding-NP stereotypical gender and reflexive pronoun gender reflects a humanlike grammatical binding domain. For each item we introduced a second profession noun, either matching or mismatching reflexive pronoun gender, in a position that linearly intervenes but is outside the reflexive's binding domain:

{\small
\setlength{\Exlabelwidth}{0.7em}
\setlength{\Exlabelsep}{0.5em}
\setlength{\SubExleftmargin}{0.7em}

\ex. \label{ex:binding-interveners}
\a. The lumberjack who is related to the soldier cut himself.\label{ex:masc-masc-masc}
\b. The lumberjack who is related to the hairdresser cut himself. \label{ex:masc-fem-masc}
\c. The lumberjack who is related to the soldier cut herself.\label{ex:masc-masc-fem}
\d. The lumberjack who is related to the hairdresser cut herself. \label{ex:masc-fem-fem}
\z.

}

\noindent
Experimental evidence suggests that humans do not consider antecedents for reflexives outside the binding domain \citep{sturt:2003,xiang-etal:2009-illusory,dillon-etal:2013-contrasting}. Likewise, if the RNN properly represents binding domains for reflexives, then only subject noun stereotypical gender, and not intervener stereotypical gender, should affect surprisal at the reflexive pronoun.

Results are in Figure~\ref{fig:gender-surprisal} (right panel).  Among conditions where intervener gender mismatches that of the reflexive pronoun (blue bars), surprisal is lower when the true antecedent matches reflexive gender ($p<.001$).  However, there is no evidence that \googlernn has learned the proper binding domain for reflexives: among conditions where true antecedent gender mismatches that of the reflexive pronoun (bars on the right), surprisal is lower when the intervener matches reflexive gender ($p<0.001$), a facilitative effect just as large as that of matching gender for the true antecedent.

\section{Negative Polarity Items}

Finally, we turn to another type of grammatical dependency: \key{negative polarity items} (NPIs).  These are items such as English ``ever'', which must be \textsc{licensed} by having an semantically negative element in a structurally appropriate context, e.g.:

{\small
\ex. \label{ex:npi-examples}
\a. No one has ever climbed that mountain.
\b. *Someone has ever climbed that mountain.
\z.

}

\noindent
We examine NPIs in English and Japanese, which (i) differ in the relative order of NPI and licensor, and (ii) have subtly different licensing domains, both of which are different than the reflexive binding domain of Section~\ref{sec:refl-pron-bind}.  Although \googlernn failed to learn reflexive binding domains, the different distributional characteristics of NPIs might well make them more learnable.

\subsection{Negative Polarity Items in English}

For present purposes, the licensing condition for English NPIs is effectively c-command by a semantically negative (downward-entailing) operator.  We investigated English NPI licensing by testing surprisal at ``any'' and ``ever'' when the potential licensor ``no'' preceded, in an appropriate position to license the NPI \ref{ex:npi-no-the}, \ref{ex:npi-no-no} and/or in a non-licensing position \ref{ex:npi-the-no}, \ref{ex:npi-no-no}:

{\small
\ex. \label{ex:npi-lisc}
\a. *The bill that the senator likes has ever found any support in the senate. \label{ex:npi-the-the}
\b. *The bill that no senator likes has ever found any support in the senate. \label{ex:npi-the-no}
\c. No bill that the senator likes has ever found any support in the senate. \label{ex:npi-no-the}
\d. No bill that no senator likes has ever found any support in the senate. \label{ex:npi-no-no}
\z.

}

\noindent
Humans are measurably surprised when encountering an unlicensed NPI as in~\ref{ex:npi-the-the} or~\ref{ex:npi-the-no}, but a non-licensing intervener as in \ref{ex:npi-the-no} elicits a ``semanticality illusion effect'' manifesting as reduced incremental processing disruption \citep{vasishth2008}.  To test whether RNNs learn the grammatical licensing condition for English NPIs, we designed 26 examples following~\ref{ex:npi-lisc} in three variants: (i) both ``ever'' and ``any'' absent; (ii) ``ever'' present, ``any'' absent; (iii) ``ever'' absent, ``any'' present.  We quantified NPI licensing effects by examining the surprisal of the NPI itself.


\begin{figure}
\centering
\includegraphics[width=.40\textwidth]{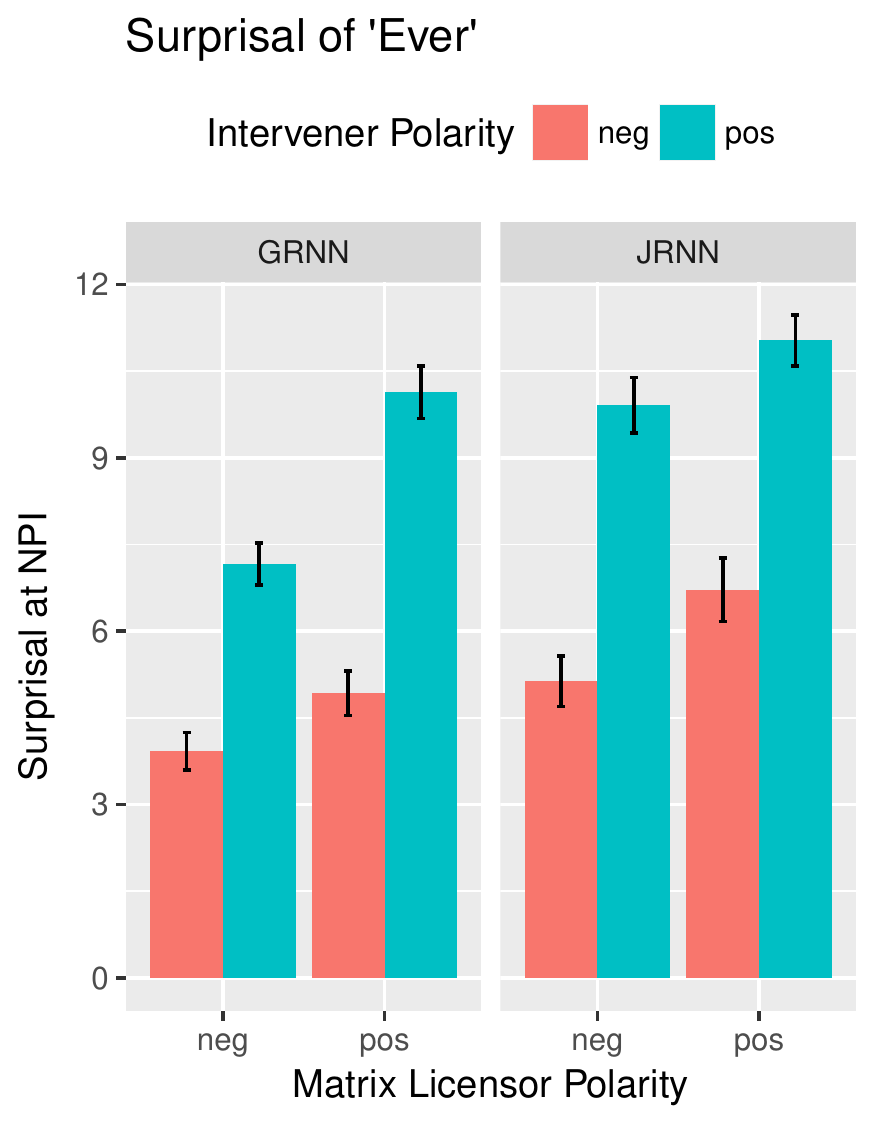}
\caption{Surprisal of NPI ``ever'' when licensed by ``no'' in a matrix clause or by a distractor ``no'' in a relative clause. Bars in each figure correspond to examples \ref{ex:npi-the-the}, \ref{ex:npi-the-no}, \ref{ex:npi-no-the}, \ref{ex:npi-no-no} from left to right.}
\label{fig:npi-eng-ever}
\end{figure}

\begin{figure}
\centering
\includegraphics[width=.40\textwidth]{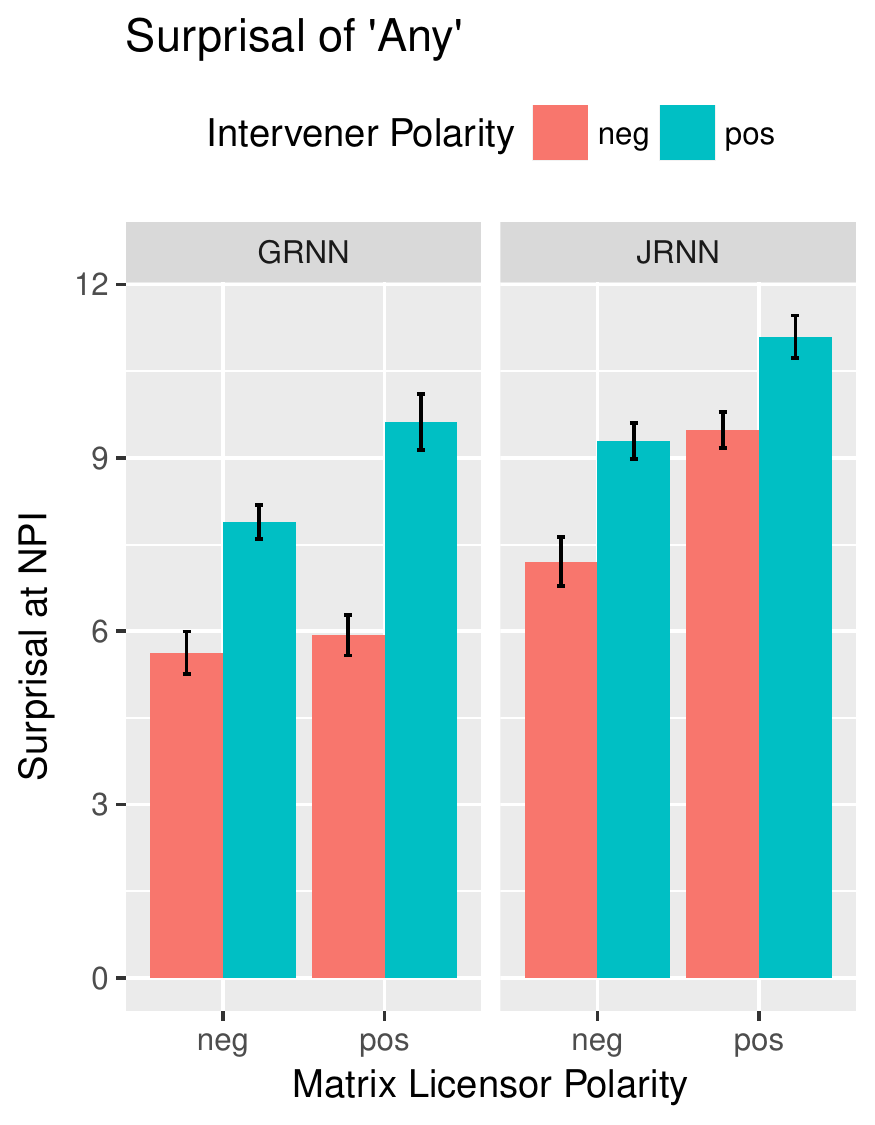}
\caption{Surprisal of NPI ``any''.}
\label{fig:npi-eng-any}
\end{figure}

Figure~\ref{fig:npi-eng-ever} shows the NPI surprisal for each of the four conditions a--d for the word ``ever'', and Figure~\ref{fig:npi-eng-any} shows the same for ``any''. The relatively high left-side bars for each condition indicate higher NPI surprisal in the absence of a grammatical licensor, as in Examples~\ref{ex:npi-the-the} and~\ref{ex:npi-the-no}. However, the relatively lower height of the red bars in both conditions indicates that surprisal is also reduced in the presence of a non-grammatical licensor in a relative cause, as in examples ~\ref{ex:npi-the-no} and ~\ref{ex:npi-no-no}.

If a model has learned the appropriate licensing conditions for English NPIs, we would expect strong surprisal reduction from ``no'' in the licensing position and zero surprisal reduction from ``no'' in the distractor position. We do find significant surprisal reduction coming from a matrix-clause ``no'' (for ``ever'', $p=0.02$ in \googlernn and $p<0.01$ in \gulordavarnn; for ``any'', $p<0.001$ in \googlernn and $p=0.03$ in \gulordavarnn), but also significant surprisal reduction coming from the distractor-position ``no'' ($p<0.001$ in both models and NPIs), indicating that the models have learned a spurious licensing relationship between a negative word embedded in a relative clause and an NPI in a higher clause, or have perhaps learned simply that any negative word licenses an NPI at any linearly following position.

\subsection{Negative Polarity Items in Japanese}
\label{sec:japanese-npi}
As described above, NPIs require negative words, but negative words do not require NPIs. In sentences and languages where the negative licensors \emph{follow} NPIs, the grammatical dependency changes to one of an obligatory upcoming event. Since such events were found to be well-represented in Section~\ref{sec:obligatory-upcoming}, sentences where negative items follow NPIs might more clearly show whether LSTMs correctly capture the grammatical dependency.  Here we consider the Japanese NPI \emph{shika}, `only', which follows this pattern:

\ex.	\label{ex:shika-basic}
	\ag.
		bus-\underline{shika}
			ko-\underline{nakat}-ta.\\
		bus-only
			come-\textsc{neg-past}\\
		`Only the bus came.'
		\label{ex:shika-basic-with-neg}
	\bg.
		*bus-\underline{shika}
			ki-ta.\\
		bus-only
			come-\textsc{past}\\
		`Only the bus came.'
		\label{ex:shika-basic-wo-neg}

In more complex sentences with embedded clauses, \emph{shika} must appear in the same clause as the negative verbs \citep[the Clausemate Condition,][]{Muraki78}: the negation must be in the main clause when \emph{shika} is not embedded \ref{ex:clausemate-condition-embedded-shika} and in the embedded clause when \emph{shika} is embedded \ref{ex:clausemate-condition-embedded-shika}.%
\footnote{
	Linguists have reported variable acceptablity of \ref{ex:clausemate-condition-embedded-shika-embedded-pos-V} (when the matrix verb is negative) depending on the grammatical role of NP-\emph{shika}. 
	\emph{Shika} in the embedded subject position
	is reported to be more acceptable
	than the direct object \citep{AoyagiIshii94,Tanaka97}
	and
	the indirect object position is the worst
	\citep{Kataoka06}.
	We did not find any experimental research
	on this issue.
}

{\small
\setlength{\Exlabelwidth}{1em}
\setlength{\Exlabelsep}{0.8em}
\setlength{\SubExleftmargin}{1.2em}

\ex. \label{ex:clausemate-condition-main-shika}
	\a.	\dots \emph{shika}\dots
			[\dots V-$\emptyset$/\textsc{neg}\dots]
			V-\textsc{neg}.
	\b.	*\dots \emph{shika}\dots
			[\dots V-$\emptyset$/\textsc{neg}\dots]
			V-$\emptyset$.

\ex. \label{ex:clausemate-condition-embedded-shika}
	\a.	\dots
			[\dots \emph{shika}\dots V-\textsc{neg}\dots]
			V-$\emptyset$/\textsc{neg}.
	\b.	*\dots
			[\dots \emph{shika}\dots V-$\emptyset$\dots]
			V-$\emptyset$/\textsc{neg}.
			\label{ex:clausemate-condition-embedded-shika-embedded-pos-V}

}

\begin{figure*}[t]
  \centering
  
\subfloat[Unembedded]{
\includegraphics[width=1.7in]{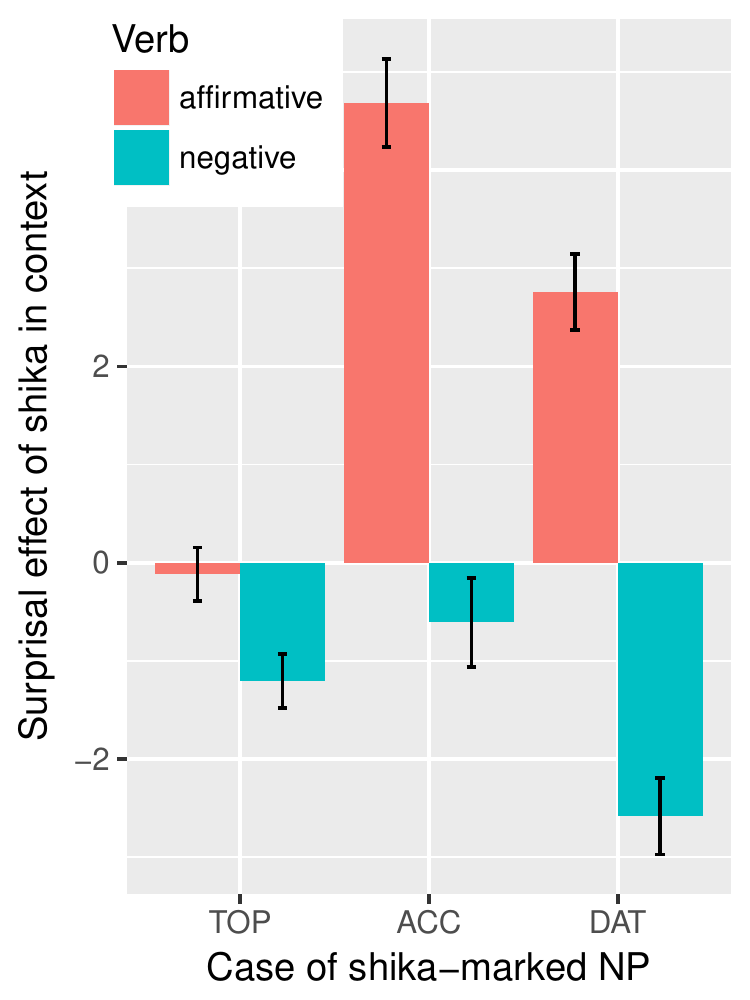}   
\label{fig:shika-unembedded} 
}
\quad
\subfloat[Embedded verb, shika in matrix clause]{
\includegraphics[width=1.7in]{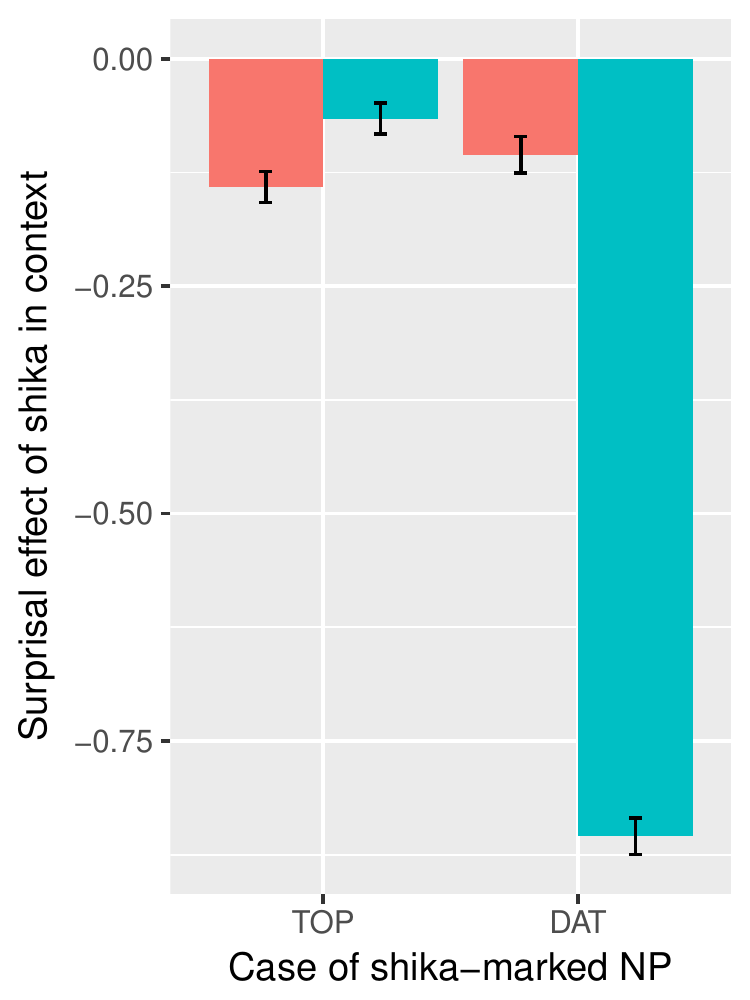}    
\label{fig:shika-matrix-verb-embedded}
}
\quad
\subfloat[Embedded verb, shika in embedded clause]{
\includegraphics[width=1.7in]{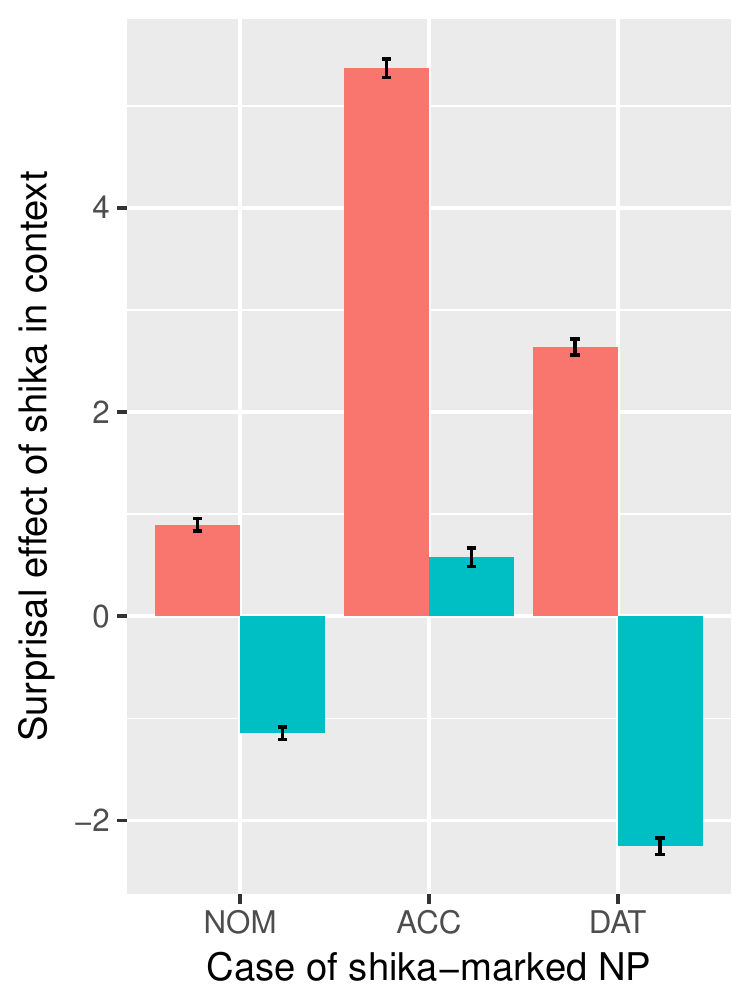}    
\label{fig:shika-embedded-verb-embedded}
}

\quad
\subfloat[Matrix verb, shika in matrix clause]{
\includegraphics[width=2in]{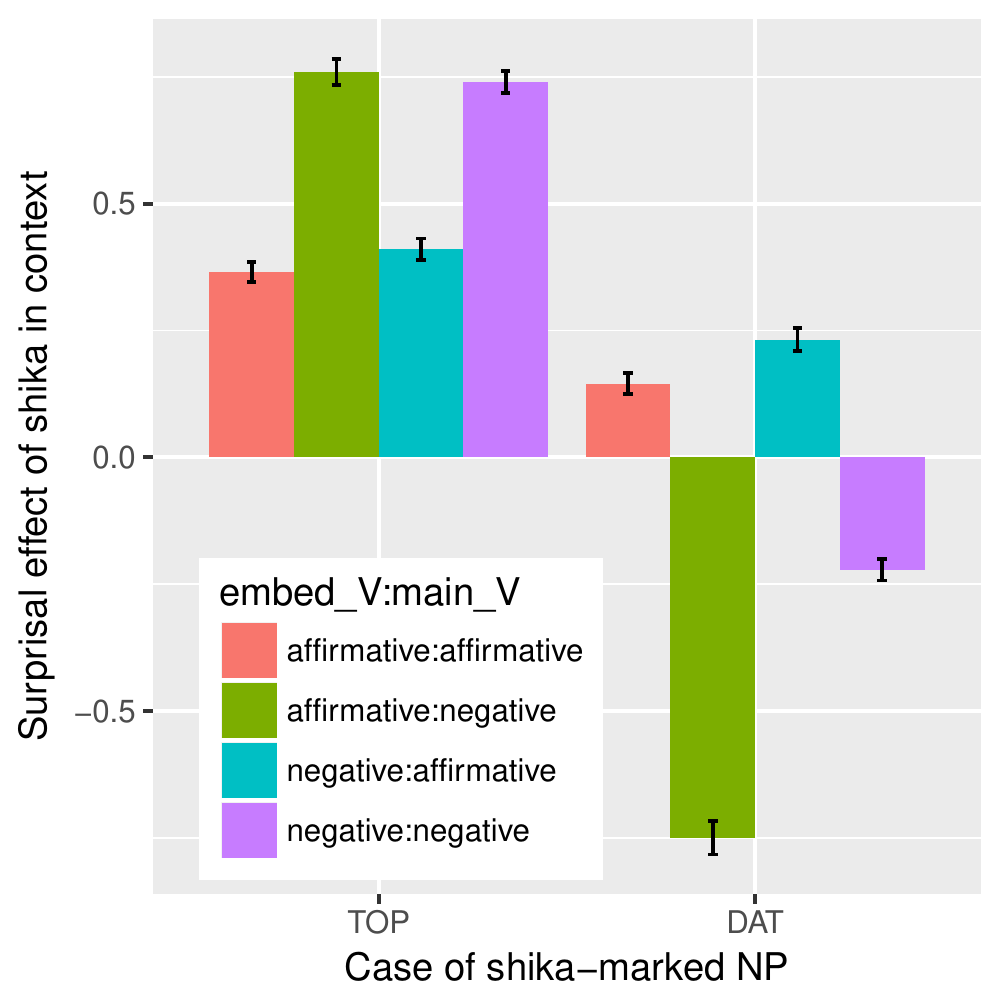}    
\label{fig:shika-matrix-verb-matrix}
}
\quad
\subfloat[Matrix verb, shika in embedded clause]{
\includegraphics[width=1.7in]{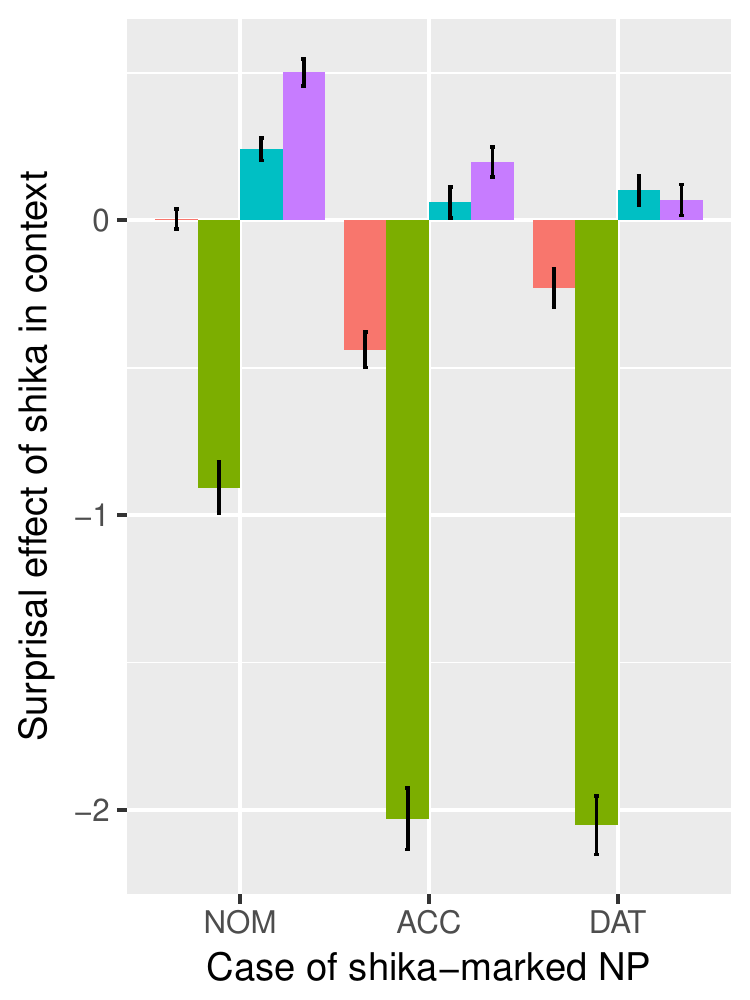}
\label{fig:shika-embedded-verb-matrix}
}

  \caption{\vspace{-0.5cm}Results for Japanese NPIs.  }
  \label{fig:japanese-npi-results}
\end{figure*}

We tested whether {\jprnn} is sensitive to these grammatical conditions by creating 83 single-clause items on the pattern of \ref{ex:shika-basic} and automatically generating 2218 items with embedded complement clauses on the patterns of \ref{ex:clausemate-condition-main-shika}--\ref{ex:clausemate-condition-embedded-shika}, varying also the case of the NP on which \emph{shika} appears. If the model has learned the proper contingency between \emph{shika} and verbal negation, then the case of NP should be irrelevant, but different cases may show different effect sizes in RNN language models because they appear with \emph{shika} with varying frequency. 

We assess how well a model has learned the \emph{shika} licensing condition by assessing the \emph{difference in surprisal} at each verb depending on whether \emph{shika} is present in a particular position in context, or absent (similar to Figure~\ref{fig:subordination-results} when we were studying subordination).  A licensing condition would manifest as \emph{shika} reducing the surprisal of a negative verb (relative to that verb's surprisal if \emph{shika} is absent) more than the surprisal of an affirmative verb; an affirmative verb in a required licensing position should show an increased in surprisal when \emph{shika} is present.

Figure \ref{fig:japanese-npi-results} shows the difference in surprisal for each condition.  Unembedded sentences (Figure~\ref{fig:shika-unembedded}) show a licensing effect for all NP cases (blue bars below 0), though we fail to get a surprisal increase for affirmative verbs when the topic is \emph{shika}-marked (the red bars are above 0 only for accusative and dative NPs).  In complex sentences where \emph{shika} is in the matrix clause, \emph{shika} on the topic NP does not lead to interpretable behavior.\footnote{There is no accusative condition in Figures~\ref{fig:shika-matrix-verb-embedded} and \ref{fig:shika-matrix-verb-matrix} because there is no verb that naturally takes both accusative object and complement clause arguments in Japanese.}
On the dative NP, \emph{shika} inappropriately leads to an expectation for negation on the embedded-clause verb (in Figure~\ref{fig:shika-matrix-verb-embedded}, the blue bar is below red bar). Furthermore, when the embedded-clause verb is affirmative, the expectation for \emph{shika} is spuriously passed on to the matrix clause verb (as shown by the negative green bar for the dative case in Figure~\ref{fig:shika-matrix-verb-matrix}). When the embedded-clause verb is negated, the expectation for a further negative verb is partially discharged (purple bar).  Finally, in complex sentences where \emph{shika} is in the embedded clause, the model generates a strong expectation for a negative embedded-clause verb (Fig.~\ref{fig:shika-embedded-verb-embedded}, blue bars below red bars), but inappropriately passes that expectation on to the matrix clause when the embedded-clause verb is affirmative (Fig.~\ref{fig:shika-embedded-verb-matrix}, green bars).

In sum, as was the case with English NPIs, our RNN clearly learns the requirement that \emph{shika} imposes for a following negative verb, but it does not learn the appropriate grammatical dependency between the NPI and the licensor.

\section{General Discussion and Conclusion}

We have applied the methods of controlled psycholinguistic experimentation to assess the evidence in contemporary RNN models for incremental syntactic state and for the proper representation of a range of grammatical dependencies. This approach builds on previous work in a similar spirit \citep{linzen2016assessing,gulordava2018colorless,wilcox2018what} and complements a variety of other approaches currently practiced \citep{shi2016string,belinkov2018evaluating,blevins2018deep,kadar2017representation,williams2018latent,ettinger2017towards,lake2017still,weber2018fine}.

In both of the English RNNs we studied we found clear evidence of incremental state syntactic representation, with important qualifications. The garden path results show that the models represent an incremental parse state inside relative clauses, and that they can partially exploit verb-form cues that indicate the onset of reduced relative clauses (Section~\ref{sec:mv-rr-verb-ambiguity}. The results on relative clause completion and subordination show strong evidence of the maintenance of expectations for obligatory upcoming material, which decays as intervening material becomes longer or more complex. 

The results on grammatical dependency show more room for improvement. None of the models tested learned the appropriate licensing conditions for reflexive pronoun binding or NPI licensing in English or Japanese. 

We believe that the psycholinguistic methodology employed in this paper provide a valuable lens on the internal representations of systems which are currently widely seen as black boxes. We have found that proper syntactic representation can emerge, but does not necessarily generalize across constructions. Future work can examine how these properties vary as a function of network architecture and objective function structure, in the pursuit of human-like syntactic competence.




\section*{Acknowledgments}

EGW would like to acknowledge support from the Mind Brain Behavior Graduate Student Grant, as well as Emmanuel Dupoux and the Cognitive Machine Learning Group at the ENS. RPL gratefully acknowledges support to his laboratory from Elemental Cognition and from the MIT-IBM Watson AI Lab. This work was supported by a GPU Grant from the NVIDIA corporation. All scripts, experimental materials, and results are available online at \url{http://github.com/Futrell/rnn_psycholinguistic_subjects}.

\bibliographystyle{emnlp_natbib}
\bibliography{everything}

\end{document}